\documentclass[conference]{IEEEtran}
\IEEEoverridecommandlockouts
\usepackage{cite}
\usepackage{amsmath,amssymb,amsfonts}
\usepackage{algorithm}
\usepackage{algpseudocode}
\usepackage{graphicx}
\usepackage{textcomp}
\usepackage{multirow}
\usepackage{soul}
\usepackage{csquotes}
\usepackage{hyperref}
\usepackage{listings}
\usepackage{lettrine}
\usepackage{array}
\usepackage{stfloats}
\usepackage{url}
\usepackage{verbatim}
\usepackage{academicons}
\usepackage{upgreek}
\usepackage{siunitx}
\usepackage{adjustbox}
\usepackage{rotating}
\usepackage[inline]{enumitem}
\usepackage{relsize}
\usepackage{longtable}
\usepackage{xcolor}
\usepackage{threeparttable}
\usepackage[caption=false]{subfig}
\usepackage{tabu}
\usepackage{color, colortbl}
\usepackage{booktabs}
\usepackage{pifont}
\usepackage{tabularx}

\begin{document}

\title{Multimodal LLM for Intelligent Transportation Systems}

\makeatletter
\newcommand{\linebreakand}{%
  \end{@IEEEauthorhalign}
  \hfill\mbox{}\par
  \mbox{}\hfill\begin{@IEEEauthorhalign}
}
\makeatother

\author{%
\IEEEauthorblockN{Dexter Le\IEEEauthorrefmark{1}}
\IEEEauthorblockA{%
\textit{Drexel University}\\
Philadelphia, USA \\
Email: dql27@drexel.edu\\
\textit{\IEEEauthorrefmark{1}Corresponding author}}

\and
\IEEEauthorblockN{Aybars Yunusoglu\IEEEauthorrefmark{1}}
\IEEEauthorblockA{\textit{Purdue University} \\
West Lafayette, USA \\
ayunusog@purdue.edu \\
\IEEEauthorrefmark{1}Corresponding author}

\and
\IEEEauthorblockN{Karn Tiwari}
\IEEEauthorblockA{%
\textit{Indian Institute of Science, Bangalore}\\
Bengaluru, India \\
karntiwari@iisc.ac.in}

\linebreakand

\IEEEauthorblockN{Murat Isik}
\IEEEauthorblockA{%
\textit{Stanford University}\\
Stanford, USA \\
misik@stanford.edu}

\and

\IEEEauthorblockN{I. Can Dikmen}
\IEEEauthorblockA{%
\textit{Temsa Research \& Development Center}\\
Adana, Turkey \\
can.dikmen@temsa.com}
}

\maketitle

\begin{abstract}
In the evolving landscape of transportation systems, integrating Large Language Models (LLMs) offers a promising frontier for advancing intelligent decision-making across various applications. This paper introduces a novel 3-dimensional framework that encapsulates the intersection of applications, machine learning methodologies, and hardware devices, particularly emphasizing the role of LLMs. Instead of using multiple machine learning algorithms, our framework uses a single, data-centric LLM architecture that can analyze time series, images, and videos. We explore how LLMs can enhance data interpretation and decision-making in transportation. We apply this LLM framework to different sensor datasets, including time-series data and visual data from sources like Oxford Radar RobotCar, D-Behavior (D-Set), nuScenes by Motional, and Comma2k19. The goal is to streamline data processing workflows, reduce the complexity of deploying multiple models, and make intelligent transportation systems more efficient and accurate. The study was conducted using state-of-the-art hardware, leveraging the computational power of AMD RTX 3060 GPUs and Intel i9-12900 processors. The experimental results demonstrate that our framework achieves an average accuracy of 91.33\%  across these datasets, with the highest accuracy observed in time-series data (92.7\%), showcasing the model's proficiency in handling sequential information essential for tasks such as motion planning and predictive maintenance. Through our exploration, we demonstrate the versatility and efficacy of LLMs in handling multimodal data within the transportation sector, ultimately providing insights into their application in real-world scenarios. Our findings align with the broader conference themes, highlighting the transformative potential of LLMs in advancing transportation technologies.
\end{abstract}


\section{Introduction}
The rise of large language models (LLM) has contributed significantly to the field of natural language processing applications. LLMs offer great potential for systems that require intelligent decision-making and can do so with great versatility. In transportation systems, making fast and accurate decisions is imperative. However, a great cost in LLMs derives from the significant cost of development and the aggregation and retention of viable data to enhance the LLM. Procedures for Data-Centric AI are thus imperative for the composition of many techniques for improving and maintaining datasets. The methodologies of Data-Centric AI range from data augmentation for data diversification, data labeling, reduction, and maintenance \cite{doi:10.1137/1.9781611977653.ch106}.

Additionally, a healthy dataset is utilized to produce a highly robust and resilient LLM that evolves alongside new data observed. In transportation systems, robustness and resilience are imperative goals for the LLM. The cost of developing LLMs is high; however, adopting these techniques can help alleviate the objectives of evolving the LLM and influencing improved continuity. Using LLMs in the transportation system can speed up accurate and intelligent responses for specific situations. A basic overview of LLMs consists of data that is initially pre-trained and tuned for instructions. These instructions are rewarded based on a reward model that features tasks. Then, the LLM initiates a prompting sequence, which is utilized for the generated response determined to be the output of the LLM \cite{naveed2024comprehensiveoverviewlargelanguage}. 


We propose a unified LLM encompassing datasets featuring time series and visuals. The novelty of a unified LLM aims to provide a framework that encapsulates the physical layer with the application layer employing machine learning algorithms. These advancements are achieved through a singular LLM architecture, which not only enhances the efficiency and accuracy of intelligent transportation systems but also facilitates real-time processing on edge devices. This architecture is particularly suitable for edge computing environments, where computational efficiency and reduced latency are critical, ensuring seamless performance in resource-constrained settings. 


The main contributions of this paper address the approach of a unified multimodal LLM framework for enhancing intelligent transport systems. Our primary contributions are:

\begin{itemize}
    \item A unified multimodal LLM framework, a novel approach that reduces the complexity of developing and increases performance for intelligent transport systems.
    \item Examination of datasets with differing data types, Time-Series, Audio, and Video; detailing various use cases with the unified architecture.
    \item  Analysis of unified multimodal LLM framework's integration into devices such as GPUs and CPUs, distinguishing performance benchmark.
\end{itemize}

The paper is organized as follows: \textbf{Section 2} discusses the current methodologies for designing an intelligent transportation system, challenges, and analysis of current learning algorithms. \textbf{Section 3} outlines the datasets used, along with the associated hardware and learning algorithms. \textbf{Section 4} describes the machine learning algorithms evaluated. \textbf{Section 5} focuses on the summary of the proposed unified multimodal LLM framework. \textbf{Section 6} discusses potential future work.

\section{Background}
Transportation is a critical component in our everyday lives, and the rising price of commodities and the demand for vehicles accelerate the need for a better transportation channel.  Additionally, transportation plays a vital role in economic development where trade can be accomplished to move goods and services \cite{ZHANG2023223}. Furthermore, transportation is significant in our contributions to environmental sustainability, safety, and community engagement \cite{hussain2024effects} \cite{DEMIREL2008279}. The emergence of Intelligent Transportation Systems (ITS) has influenced the manners of transportation by positively improving existing issues, including but not limited to: traffic safety, cost efficiency, comfort, and speed \cite{waqar2023evaluation} \cite{ELASSY2024100252}. The applications of ITS vary from autonomously driving vehicles to traffic management.  Where the predominant rise of IoT devices has accelerated the employment of ITS \cite{khalid2019fog}.  Similarly, the emergence of cloud computing has further influenced the development of IoT devices to integrate with ITS to produce a highly robust and resilient system \cite{Mnyakin_2023}.  Transportation is essential to everyday life, and safety, where fostering a sustainable model should be paramount.

Despite the emergence of ITS, transportation still faces many challenges that impact livelihoods.  The challenges that surround transportation range from environment to health, safety, privacy, and accessibility.  Environmental challenges stem from the byproducts of transportation, the act of transporting or manufacturing capable products.  A tremendous environmental challenge is the emission to the atmosphere of carbon dioxide as a result of fuel combustion \cite{COLVILE20011537}. This environmental challenge also affects the health of others, as mass emission reduces air quality and can be a factor in life-long diseases.  Safety plays a crtical role in transportation systems, where some of the most significant challenges are general safety, sustainability, and autonomous transportation \cite{KAEWUNRUEN2016}.

\vspace{5pt}


\vspace{5pt}


\vspace{5pt}

\begin{figure}[h]
    \centering
    \graphicspath{ {D:\Stack} }
    \includegraphics[width=0.45\textwidth]{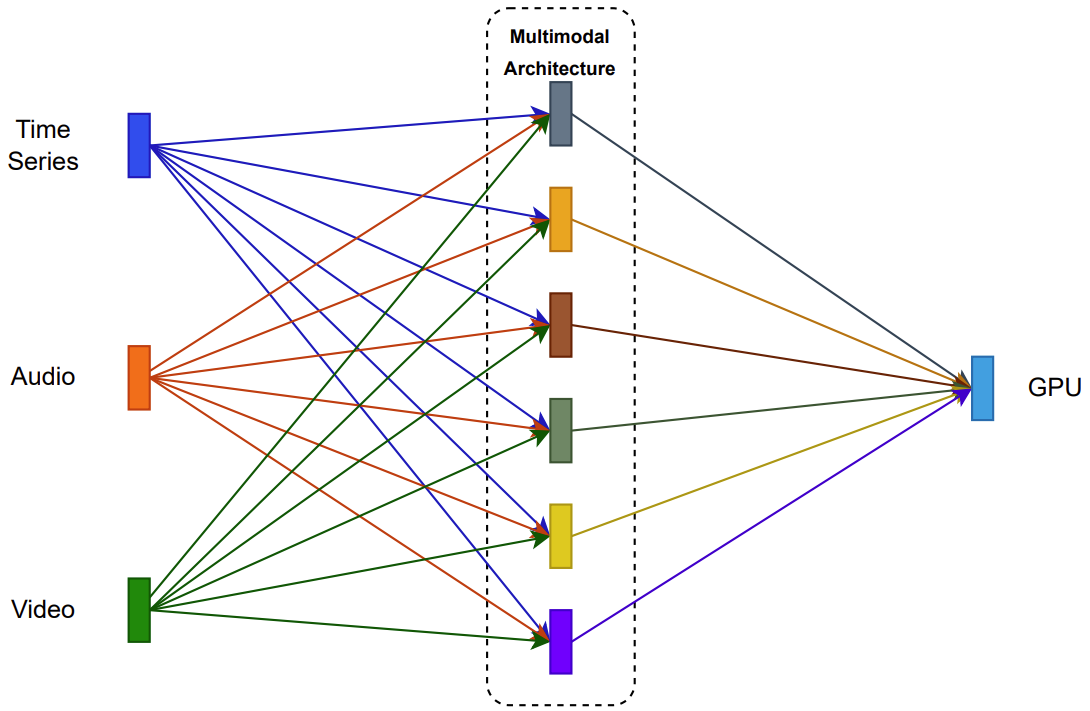}
    \caption{\centering Our Framework Depicted Across Three Dimensions: Data, Models, and Hardware.}
    \label{Figure 4}
\end{figure}

\autoref{Figure 4} illustrates the interactions within the Multimodal LLM Framework span three key dimensions: Modality, Models, and Hardware. Each modality, including time series, audio, and visual data, is represented as an input to the network, ensuring the framework effectively processes and analyzes diverse data formats across these dimensions. Then, each node applies to subsequent learning algorithms in the physical layer. Each node in the network and its interactions with other layers display the challenges of the Multimodal LLM Framework and the tasks to accomplish—traditional scalability results in an exponential increase in dependencies and complexities. The approach of the Multimodal LLM Framework aims to reduce scaling complexities while improving performance and abstracting interactions of the input layer to the physical. 

\section{Methods}

\subsection{Dataset}
The dataset utilized in this study is derived from time-series sensor data and encompasses a wide range of parameters related to vehicle performance, control systems, and environmental conditions. This rich dataset is designed to support analyzing intelligent transportation systems and their interactions with the surrounding environment. 

\begin{table}[h]
    \centering
    \footnotesize 
    \caption{Overview of different data types in the dataset.}
    \resizebox{0.45\textwidth}{!}{
    \begin{tabularx}{0.5\textwidth}{lXX} 
        \toprule
        \textbf{Data Type} & \textbf{Description} & \textbf{Applications} \\
        \midrule
        Time-Series Data & Continuous sensor readings over time, including vehicle speed, tire pressure, engine torque, etc. & Analyzing dynamic behavior, fault detection, predictive maintenance \\
        Audio Data & Sound recordings from the vehicle’s surroundings or internal systems. & Noise analysis, engine sound and environmental conditions assessment \\
        Video Data & Recordings from onboard cameras capturing the vehicle’s environment. & Object detection, lane-keeping assistance, environmental monitoring \\
        \bottomrule
    \end{tabularx}}
    \label{tab:dataset_overview}
\end{table}

\autoref{tab:dataset_overview} shows the datasets comprise three primary types of data; time-series, audio, and video each serving distinct purposes in vehicular analysis and applications. Time-series data includes continuous sensor readings over time, such as vehicle speed, tire pressure, and engine torque, and is essential for analyzing dynamic behavior, fault detection, and predictive maintenance. The audio data consists of sound recordings taken from the vehicle's surroundings and internal systems, which can be used for noise analysis, engine sound evaluation, and evaluation of the vehicle's environment. In video data, onboard cameras record the vehicle's environment, which is used to detect objects, assist with lane-keeping, and monitor the environment. We intend to evaluate LLMs' ability to process and interpret multimodal sensor data by utilizing this dataset. Using LLMs, this research aims to improve understanding of complex vehicle-environment interactions, ultimately leading to advances in intelligent transportation.

\begin{figure}[h]
    \centering
    \graphicspath{ {D:\Stack} }
    \includegraphics[width=0.5\textwidth]{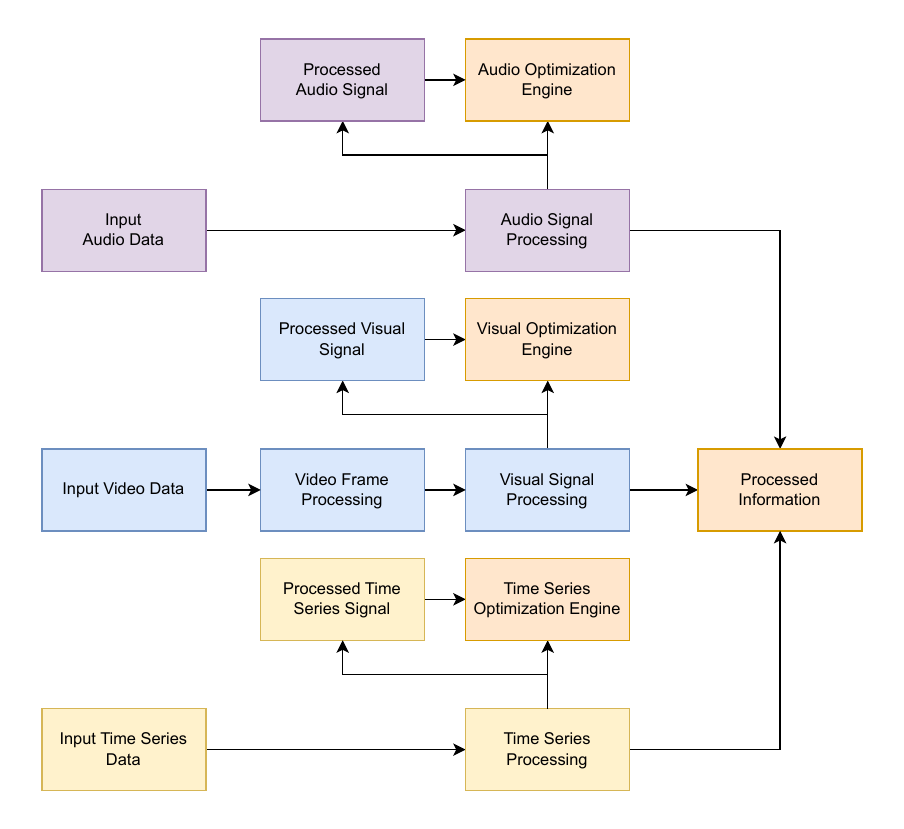}
    \caption{Sensor Processing Diagram.}
    \label{Figure 2}
\end{figure}

\vspace{5pt}

\autoref{Figure 2} illustrates the sensor processing of the Multimodal LLM Framework, which integrates time-series, audio, and visual data. Data inputs, such as time series and audio, are processed by their optimization engines in a loopback mechanism. The input video frames are preprocessed first for visual data. A loopback process is used by the visual optimization engine to process the visual frame after preprocessing. For their tasks, the optimization engines use the AdamW optimizer and continuous learning loops. Information that has been processed is the output of each signal processing module. The framework provides a scalable solution that maintains efficiency without increasing complexity while supporting various data formats.

\subsection{Overview of the Proposed Framework}
The proposed framework integrates time series, audio, and visual data processing with the most advanced developments, the proposed framework illustrates the versatility of converter-based architectures across a variety of data formats. The framework consists of three main components: time series analysis, audio classification, and visual data processing. In order to classify structured tabular data, the framework implements a Converter architecture for time series data, specifically BERT. The features in the tabular data are converted into a single text string, allowing the BERT tokenizer to process the data and capture complex relationships in the dataset \cite{devlin2019bertpretrainingdeepbidirectional}. The architecture comprises a pre-trained BERT model followed by a fine-tuned classification head on the target dataset. Training is optimized using the AdamW optimizer, and the model's performance is evaluated using accuracy metrics \cite{loshchilov2019decoupledweightdecayregularization}. This approach enables the model to handle complex time series data effectively, providing robust performance on unseen data. The framework's audio classification component leverages a pre-trained Wav2Vec2 model, fine-tuned on a specific audio dataset for classifying environmental audio files \cite{baevski2020wav2vec20frameworkselfsupervised}. The process begins with data preprocessing, where Wav2Vec2Processor converts the raw audio waveforms into a format suitable for the Wav2Vec2 model, including padding and normalization. An AudioDataset class is implemented to manage the audio data appropriately, ensuring correct label assignment and necessary transformations. Audio sequences vary in length and are processed by a custom sorting function that pads the sequences in each batch to the maximum sequence length, ensuring consistent input sizes for the model during training. The model is then fine-tuned using the cross-entropy loss and AdamW optimizer, and evaluation is based on validation accuracy and loss metrics.

\vspace{5pt}

Video sequences are processed in the visual data processing component by extracting frames and generating textual descriptions. The descriptions are based on image captioning models like BLIP and CLIP. Based on the generated descriptions, a language model, like T5, refines or transforms the text based on specific natural language processing (NLP) tasks, like translation or summarization. This process enables visual and textual data integration, leveraging transfer learning to minimize the need for task-specific training. The entire visual processing pipeline is implemented using Hugging Face Transformers library, ensuring flexibility and compatibility with different pre-trained models.

\vspace{5pt}

The multimodal framework shows how LLMs can be applied to a variety of data types, such as time series, audio, and visual, illustrating the model's adaptability and effectiveness. In addition to leveraging the contextual understanding provided by pre-trained models, the framework achieves high performance with minimal computational resources by transforming non-textual data into formats suitable for NLP models. Multimodal systems are particularly useful for complex tasks that take advantage of rich, contextual representations provided by transformers. This makes the framework applicable to a variety of domains, including predictive maintenance, audio event detection, and video analysis.

\begin{figure*}[h]
    \graphicspath{ {D:\Stack} }
    \center 
    \includegraphics[width=0.9\textwidth]{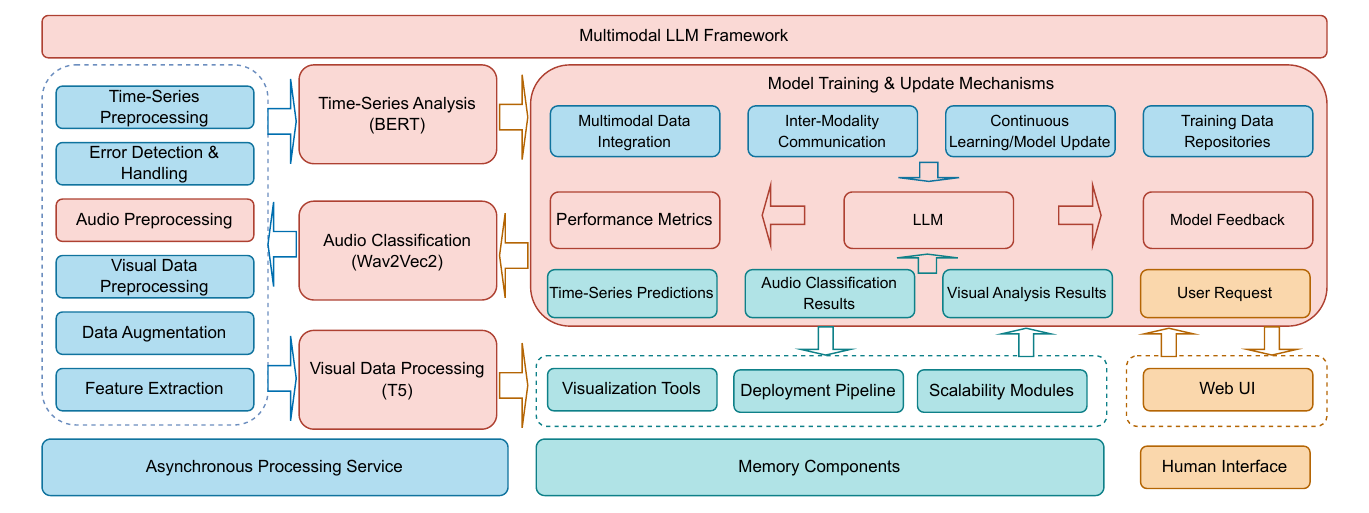}
    \caption{Block Diagram of Implementation.}
    \label{Figure 3}
\end{figure*}

\vspace{5pt}

\autoref{Figure 3} shows that the proposed Multimodal LLM Framework integrates time-series, audio, and visual data processing using transformer-based models. It begins with a Preprocessing Layer that includes Time-Series, Audio, Visual Data Preprocessing modules, Error Detection and Handling, along with Feature Extraction. Then, the data is processed using transformers that are specific to the modality: Time-Series Analysis (BERT), Audio Classification (Wav2Vec2), and Visual Data Processing (T5). Multimodal Data Integration is at the core of the framework, which provides seamless intermodal communication and incorporates an LLM model feedback loop for continuous learning and model updates. Models are trained and updated through the framework, which is guided by performance metrics and supported by training data repositories. An Asynchronous Processing Service manages tasks and facilitates user interaction with a Web UI and Human Interface. Integration of models into production environments is ensured by the Deployment Pipeline. The framework provides sophisticated analysis and predictions across time series, audio, and visual modalities, making it a scalable and flexible multimodal data processing solution.

\subsection{Hardware Implementation}
Our algorithms were implemented using Python on CPUs and GPUs. The study was carried out by leveraging the computational power of NVIDIA's GeForce RTX 3060 GPU and Intel's Core i9 12900H CPU, optimized for different tasks, ensuring efficient execution of our implementations.

\section{Evaluation}
The evaluation of the proposed framework is centered on assessing the performance and computational efficiency of the integrated LLMs when applied to multimodal data in intelligent transportation systems. The framework's performance was demonstrated across varying data types, including time series, audio, and video data depicting different aspects of vehicle and environmental conditions.

\begin{table}[h]
    \centering
    \small
    \setlength{\tabcolsep}{2pt} 
    \renewcommand{\arraystretch}{1.3} 
    \caption{Evaluation Results for Multimodal LLM in Intelligent Transportation Systems}
    \label{table:table_2}
    \resizebox{0.4\textwidth}{!}{
    \begin{tabular}{lcccc}
        \toprule
        & \textbf{Accuracy} & \textbf{MAC (GOP)} & \textbf{Task} & \textbf{Latency (ms)} \\ 
        \midrule
        \textbf{Time-Series} & 94.48\% & 1.8 & Classification & 11.5 \\ 
        \textbf{Audio} & 92.80\% & 2.7 & Classification & 13.1 \\ 
        \textbf{Video} & 88.73\% & 4.5 & Captioning & 13.5 \\ 
        \bottomrule
    \end{tabular}}
    \label{tab:comparison1}
\end{table}

\autoref{tab:comparison1} shows the performance of a multimodal LLM across three different modalities: time series, audio, and video. The model showcases an impressive accuracy of 94.48\%  for time-series classification with a latency of 11.5 ms and a computational complexity of 1.8 GOPs. Audio classification follows with an accuracy of 92.80\%, a latency of 13.1 ms, and a computational demand of 2.7 GOPs. Video processing, tasked with captioning, presents a lower accuracy of 88.73\% but requires the highest computational effort of 4.5 GOPs, with a latency to audio at 13.5 ms.


\vspace{5pt}


\section{Conclusion} 

The results demonstrate that the proposed multimodal LLM framework balances accuracy and computational efficiency across various data modalities, making it highly suitable for autonomous driving applications. The framework excels in processing time-series data, which is crucial for motion planning and real-time decision-making in transportation systems. While audio and video data performance showed room for improvement, these results highlight potential areas for further optimization, particularly in handling more complex and high-dimensional inputs. The latency results confirm that the model is capable of real-time processing, even with computational constraints, making it viable for deployment in real-world intelligent transportation systems. Transfer learning and task-specific fine-tuning also allowed the model to achieve robust performance without excessive computational demands. The proposed framework offers a robust and efficient solution for integrating LLMs into autonomous driving and other transportation-related tasks, providing high accuracy and computational efficiency.

\section{Future Work}


The LLM framework features a multimodal approach that encapsulates varying datasets for an intelligent transportation system. Potential fields of improvement could be accomplished by testing the LLM framework with another dataset. Utilizing other datasets can help improve the LLM framework and make cross-validation possible. Similarly, applying techniques for augmenting the dataset can validate any potential overfitting in the results. Another challenge in multimodal systems is the integration of diverse data formats (time series, audio, visual) into a shared latent space. In future work, t-Distributed Stochastic Neighbor Embedding (t-SNE) or similar dimensionality reduction techniques can be employed to visualize and optimize the learned latent representations. These techniques can help assess whether the LLM framework is effectively grouping semantically similar data points across different modalities. Exploring the prospect of varying extra knowledge prompting concerning sample quantity could improve the LLM framework performance \cite{qi2023limitation}. Additionally, explorations of visual and text prompting improvements could range from reliance on linguistic biases to crucial information in the text. 

\section{Acknowledgment}
We acknowledge the Temsa Research R\&D Center for their generous financial support and the reviewers for their invaluable insights and suggestions that significantly contributed to the enhancement of our paper.

\bibliographystyle{IEEEtran}
\bibliography{external}

\end{document}